\documentclass[10pt,twocolumn,letterpaper]{article}

\usepackage{iccv}
\usepackage{times}
\usepackage{epsfig}
\usepackage{graphicx}
\usepackage{amsmath}
\usepackage{amssymb}
\usepackage{multirow}
\usepackage{enumitem}
\usepackage{caption}
\usepackage{subcaption}
\usepackage{xcolor}
\usepackage{amsthm}
\usepackage[ruled,vlined]{algorithm2e}
\usepackage{tabularx,booktabs}
\usepackage{hhline}
\usepackage{pifont}
\usepackage{diagbox}
\usepackage{makecell}

\usepackage[pagebackref=true,breaklinks=true,letterpaper=true,colorlinks,bookmarks=false]{hyperref}

\iccvfinalcopy 

\def\iccvPaperID{8101} 

\ificcvfinal\fi

\newcommand{\VE}{UVO}

\begin{document}

\makeatletter
\DeclareRobustCommand\onedot{\futurelet\@let@token\@onedot}
\def\@onedot{\ifx\@let@token.\else.\null\fi\xspace}
\def\eg{\emph{e.g}\onedot} \def\Eg{\emph{E.g}\onedot}
\def\ie{\emph{i.e}\onedot} \def\Ie{\emph{I.e}\onedot}
\def\cf{\emph{c.f}\onedot} \def\Cf{\emph{C.f}\onedot}
\def\etc{\emph{etc}\onedot} \def\vs{\emph{vs}\onedot}
\def\wrt{w.r.t\onedot} \def\dof{d.o.f\onedot}
\def\etal{\emph{et al}\onedot}
\makeatother

\title{Unidentified Video Objects: A Benchmark for Dense, Open-World Segmentation}

\author{%
  Weiyao Wang, Matt Feiszli, Heng Wang, Du Tran \\
  Facebook AI Research (FAIR)\\
  \texttt{\{weiyaowang,mdf,hengwang,trandu\}@fb.com} \\
}

\maketitle
\ificcvfinal\thispagestyle{empty}\fi


\begin{abstract}
Current state-of-the-art object detection and segmentation methods work well under the closed-world assumption. This closed-world setting assumes that the list of object categories is available during training and deployment. However, many real-world applications require detecting or segmenting novel objects, i.e., object categories never seen during training. In this paper, we present, \VE{} (\textbf{U}nidentified \textbf{V}ideo \textbf{O}bjects), a new benchmark for open-world class-agnostic object segmentation in videos. Besides shifting the problem focus to the open-world setup, \VE{} is significantly larger, providing approximately 8 times more videos compared with DAVIS, and 7 times more mask (instance) annotations per video compared with YouTube-VOS and YouTube-VIS. \VE{} is also more challenging as it includes many videos with crowded scenes and complex background motions. We demonstrated that \VE{} can be used for other applications, such as object tracking and super-voxel segmentation, besides open-world object segmentation. We believe that \VE{} is a versatile testbed for researchers to develop novel approaches for open-world class-agnostic object segmentation, and inspires new research directions towards a more comprehensive video understanding beyond classification and detection.


\end{abstract}

\section{Introduction}
\label{sec:intro}

\begin{figure}
\captionsetup{font=footnotesize}
{\small 
 \begin{subfigure}[b]{0.3\linewidth}
    {\includegraphics[width=\linewidth]{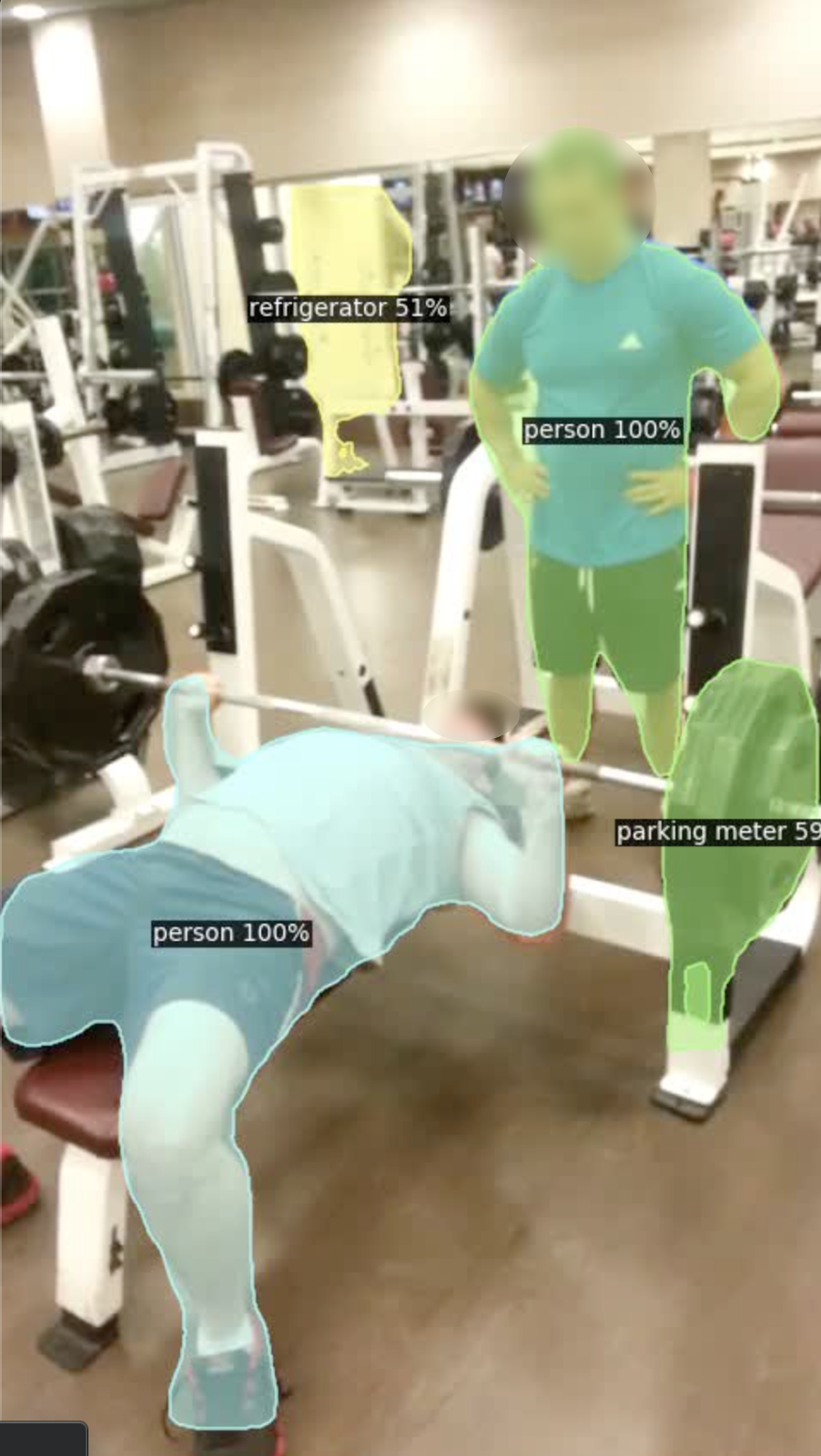}}
    \caption{}
 \end{subfigure}
 \begin{subfigure}[b]{0.3\linewidth}
    {\includegraphics[width=\linewidth]{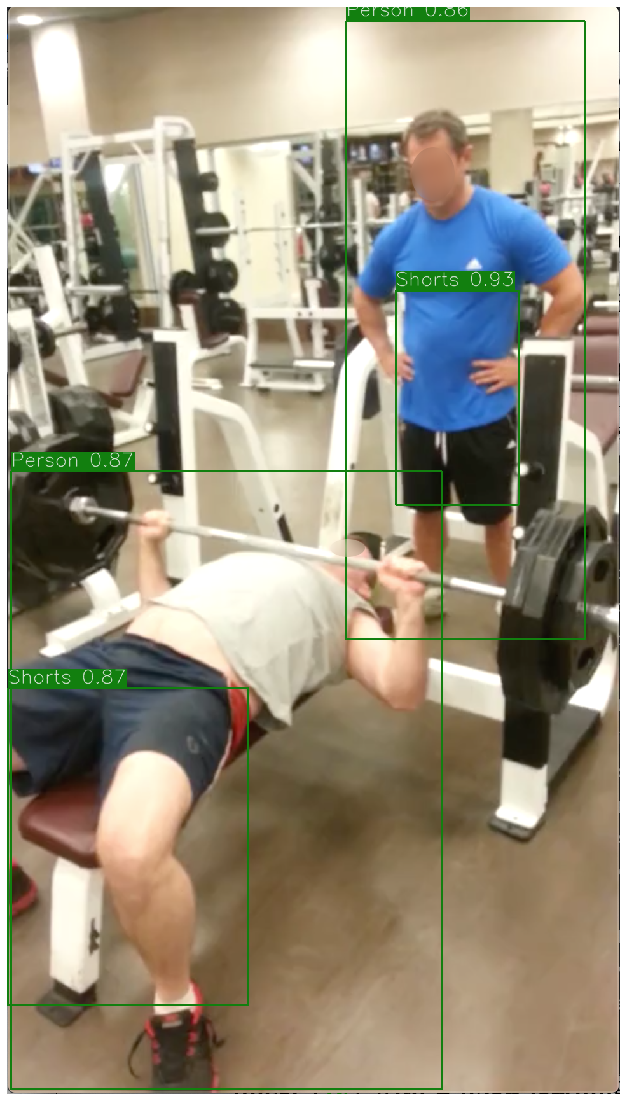}}
    \caption{}
 \end{subfigure}
 \begin{subfigure}[b]{0.3\linewidth}
    {\includegraphics[width=\linewidth]{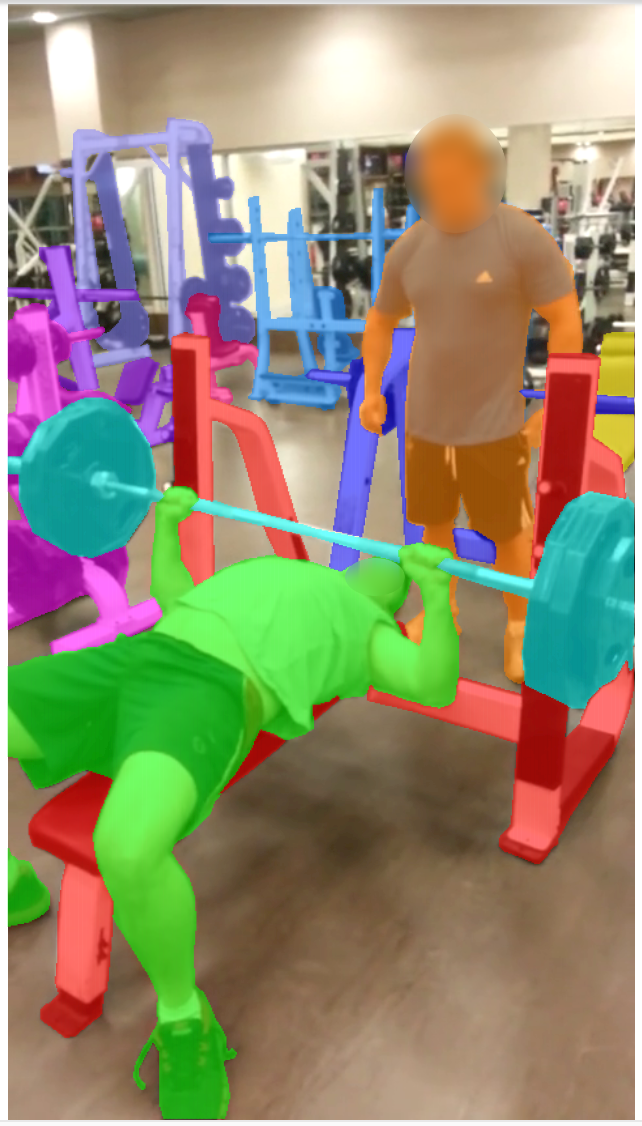}}
    \caption{}
 \end{subfigure}
 \vspace{-8pt}
 \caption{{\bf State-of-the-art object detection/segmentation methods do not work well in open-world settings.} We evaluated (a) Mask R-CNN trained on COCO and (b) Google AI cloud API on Kinetics-400 videos, and found both methods fail to segment many objects that have not been seen in training.
 (c) Real-world applications require segmenting all objects that appear in the videos, even unseen objects. 
 Mask R-CNN works well only on predefined categories and fails to recognize objects (\eg, barbell) or confuses non-object with objects in the taxonomy (refrigerator). Google cloud object detector offers stronger detection results but still misses all gym equipment in the background. (c) \VE{} is designed to 
 detect/segment all objects regardless of the categories and beyond.
 }
 \label{fig:compare_vis}
 }
\end{figure}

During daily activities, humans routinely encounter novel objects, \eg, unfamiliar birds or unknown flowers; despite this unfamiliarity, people have no problem perceiving them as distinct object instances. Even cinematic examples like UFOs will be identified as independent things. Many real-world applications such as object search~\cite{7328738,4270197}, instance registration~\cite{Zhang_2017_ICCV}, human-object interaction modeling~\cite{GkioxariHOI} and human activity understanding~\cite{Damen2018EPICKITCHENS} require such open-world prediction abilities, \eg, exhaustively detecting or segmenting objects (both known and unknown), to accomplish their tasks. Open-world is also the natural setting for applications like embodied AI (\eg, robotics, autonomous driving) and augmented-reality assistants, which will encounter novel situations regularly.


In contrast, the current state-of-the-art is designed for closed-world detection and segmentation. Current Mask R-CNN~\cite{8237584}, when trained on COCO~\cite{Lin2014MicrosoftCC} with a closed-world taxonomy of 80 object categories, cannot segment novel objects (see Figure~\ref{fig:compare_vis}a); the model can only be expected to segment the classes it is trained for. Similarly, an industry publicly-available object detector (capable of detecting 550+ objects)~\cite{G-Cloud} still cannot detect many objects (see Figure~\ref{fig:compare_vis}b). From a methodology perspective, open-world object segmentation is a challenging problem due to its open-taxonomy nature. 
Though extensive research has been done to develop either supervised top-down approaches~\cite{8237584,FasterRCNN,7780460} or unsupervised bottom-up approaches~\cite{SelectiveSearch,868688,GBH} for object detection/segmentation in the closed-world setting, 
we find that simple modifications of existing approaches does not work well in the open-world segmentation problem. On one hand, adopting a top-down approach, \eg, Mask R-CNN, to class-agnostic object segmentation by replacing its multi-class loss with a binary foreground versus background loss performs poorly on unseen classes (as shown in Table~\ref{tab:open_world_result}). This is because the top-down approaches are strongly biased toward contextual cues from seen classes~\cite{singh-cvpr2020}. On the other hand, unsupervised approaches, \eg, GBH~\cite{GBH}, rely on pixel grouping using local cues, \eg, colors and/or motions, which have no notion about semantic object boundaries, thus also perform poorly (see Figure~\ref{fig:gbh_videoentity}). We believe that open-world object segmentation is a challenging yet meaningful 
problem that needs to be addressed by approaches that are conceptually different from existing methods.

Open-world object segmentation also provides opportunities for long video modeling and more complex prediction tasks. Current video understanding approaches~\cite{Tran15,I3D,Tran18,slowfast,CSN19} cannot scale well to long videos due to GPU memory constraints and are not designed for complex prediction tasks beyond classification and detection. Grouping pixels into semantic entities (
including unknown classes) will provide plausible alternatives for long-term video modeling and flexible prediction tasks, \eg, reasoning about objects and their interactions, even when their types are unknown, possibly by applying graph convolutional networks~\cite{XiaolongWangGCN18,Wang_2021_WACV}, knowledge graphs~\cite{Yuan2017EndtoEndVC,ghosh2020knowledge}, or attention-based models~\cite{XiaolongWang18,vision_transformer} on top of entities instead of pixel-based CNN which is memory-intensive.


Due to its broad applications, open-world object segmentation is an important problem to study. Current datasets are often constructed with predefined taxonomy in a closed-world manner. Removing object labels in existing datasets will not make them 
suitable for open-world segmentation: to evaluate detector performances in open-world, a dataset needs to contain exhaustive annotations on all objects; otherwise, a model detecting un-annotated objects is not rewarded and can even be penalized.
Ideally, the dataset should be annotated in a bottom-up fashion: annotators watch videos, spot and mask all objects. To our knowledge, no existing dataset provides such exhaustive annotations at scale for videos or images~\footnote{LVIS~\cite{8954457} is a federated dataset of multiple small shards; each contains exhaustive annotations \wrt a subset of classes within the shard but not inter-shard: \eg, ``person'' is not annotated in many images.}. Bottom-up annotation pipelines are absent in videos and are rare in images~\footnote{LVIS is bottom-up in federated setting; ADE20k~\cite{8100027} is smaller scale.}. In this paper, we make the first step forward in addressing this problem by constructing a new benchmark for open-world object segmentation, and providing a comprehensive set of baselines with in-depth analysis to understand the benchmark and problem. Our contributions are:
\begin{itemize}[noitemsep]
\item A method for constructing open-world object segmentation datasets using object interpolation and tracking, which is {\bf 4x} more efficient than the baseline.
\item We introduce, \textbf{U}nidentified \textbf{V}ideo \textbf{O}bjects (UVO), a new benchmark for open-world object segmentation. Besides shifting the focus to the open-world setup, \VE{} has approximately {\bf 8x} more videos compared with DAVIS~\cite{7780454}, and {\bf 7x} more mask (instance) annotations per video compared with YouTube-VOS~\cite{YouTubeVOS} and YouTube-VIS~\cite{YouTubeVIS}.
\item We provide a comprehensive set of baselines to understand our proposed tasks and benchmark. We believe that \VE{} is a versatile testbed for open-world object segmentation and will inspire new research directions toward more complex video understanding tasks beyond classification and detection.
\end{itemize}


\section{Related Work}
\label{sec:related_work}

\textbf{Open-world object recognition and detection}. Open-world problems have been studied in the context of recognition~\cite{bendale2015towards,liu2019large}: given a closed-world training dataset, how to actively identify new object categories? 
To handle novel objects, previous works explicitly differentiate unknown from known, such as by spotting outliers in the embedding space. There are also previous studies on open-world object mask prediction. Pinheiro~\etal~\cite{DeepMask15} trained a class-agnostic mask predictor. Hu~\etal~\cite{8578543} proposed an approach for predicting unknown object masks using known object bounding boxes. More recently, Jaiswal~\etal~\cite{Jaiswal_2021_WACV} proposed an adversarial framework to learn out-of-taxonomy objects. Dhamija~\etal~\cite{9093355} discussed the difficulties in open-world object detection. Despite the aforementioned works, we still lack of a dedicated dataset on this topic. This inspires us to create \VE{} to facilitate more extensive research efforts on open-world object detection and segmentation.



\textbf{Related datasets}. Object detection and segmentation have been the focus of computer vision for the past decades. 
Much of the progress we have achieved so far is built upon pioneering datasets: BSDS~\cite{MartinFTM01}, Caltech101~\cite{1597116}, PASCAL-VOC~\cite{PascalVOC}, COCO~\cite{Lin2014MicrosoftCC}, ADE20k~\cite{8100027}, LVIS~\cite{8954457}, \etc. Many recent datasets extend the task from image to video: DAVIS~\cite{7780454}, YouTube-VOS (YTVOS)~\cite{YouTubeVOS}, MOTS~\cite{Voigtlaender_2019_CVPR}, YouTube-VIS (YTVIS)~\cite{YouTubeVIS}, TAO~\cite{10.1007/978-3-030-58558-7_26}. \VE{} is inspired by the aforementioned datasets, but have several key features differentiating itself from previous ones.
Existing datasets typically rely on fixed taxonomies, such as ``salient objects'' (DAVIS) or a list of object categories (YTVOS/YTVIS) (see Table~\ref{tab:dataset_comparison}). On the contrary, \VE{} is taxonomy-free, and provides comprehensive annotations for 
all the objects in the open world. As a result, our dataset contains significantly more instances per video compared to previous video object segmentation datasets.

\section{Open-World Object Segmentation}
\label{sec:dataset}

Compared with traditional object segmentation task, the open-world setup requires to segment all the entities or objects \emph{class-agnostically} and \emph{exhaustively}. These requirements ensure the model to detect unseen categories during testing, and learn more complete representations of the visual world. We detail the process of building the dataset and provide in-depth analysis about its characteristics in the following sections.

\subsection{An overview of UVO}

\begin{figure}
\captionsetup{font=footnotesize}
{\small
\begin{center}
\includegraphics[width=0.8\linewidth]{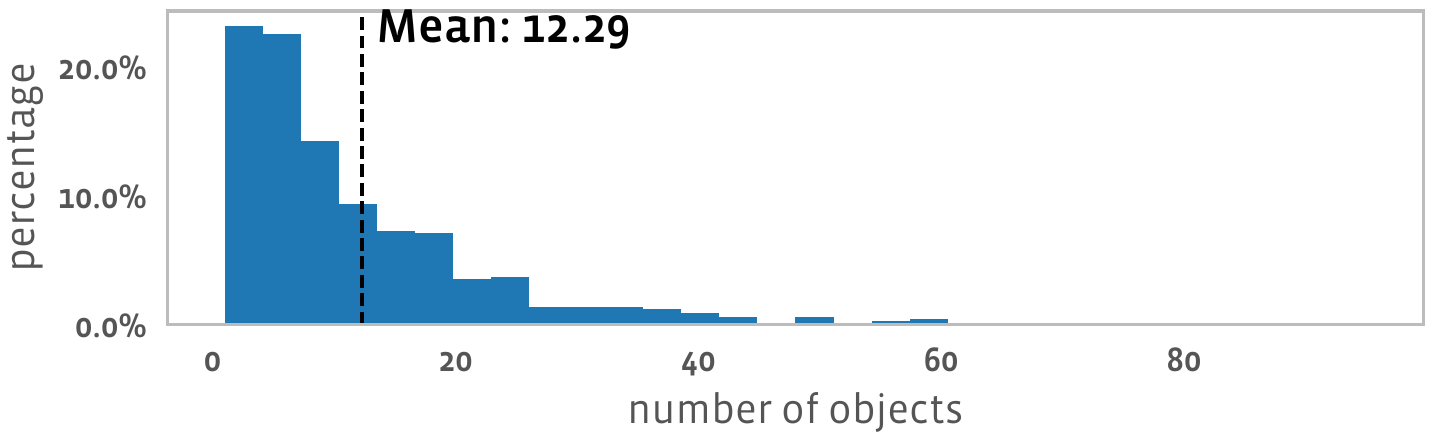}
\vspace{-8pt}
\caption{{\bf Distribution of the number of objects per video}. The distribution is long-tail, with a mean of 12.29 and a median of 8. In some extreme cases, we observe nearly 100 object instances in a videos. This is close to real world distribution and more suitable for open-world applications.}
\label{fig:object_distribution}
\end{center}}
\end{figure}

We introduce a new benchmark: the \emph{\VE{}} dataset for open-world object segmentation. \VE{} contains real-world videos, with dense and exhaustive object mask annotations. Dense object segmentations are extremely costly to annotate, and no existing dataset provides such annotations at a large scale. Real-world videos often contain dozens of object instances (see Figure~\ref{fig:object_distribution}), and it takes a trained annotator 45 minutes per frame to densely annotate all the objects and link the objects across frames. 
We build a semi-automatic approach to accelerate the process and save the annotation cost. 
In this section, we provide an overview of UVO. 

\begin{table}[!h]
\captionsetup{font=footnotesize}
    \centering
    {\small
    	\begin{tabular}{|c|c|c|c|c|}
        	\hline
        	Dataset & \makecell{Videos \\ / Frames} & Taxonomy & \makecell{Objects \\ per video} & \makecell{Ann. \\ fps} \\
        	\hline
        	DAVIS~\cite{DAVIS2017} & 150/11k & ``salient" & 2.99 & 24 \\
        	\hline
        	YTVOS~\cite{YouTubeVOS} & 4453/120k & 94 classes & 1.64 & 6 \\
        	\hline
        	YTVIS~\cite{YouTubeVIS} & 2883/78k & 40 classes & 1.68 & 6 \\
        	\hline
        	\VE{} & 1200/108k & open-world & 12.29 & 30 \\
        	\hline
    	\end{tabular}
	}
	\vspace{-6pt}
	\caption{\textbf{Comparison with current datasets.} \VE{} is the largest compared with previous datasets in terms of the number of frames. \VE{} has no predefined taxonomy, but all objects are exhaustively annotated, resulting in a significantly larger number of objects per video. 
	}
	\label{tab:dataset_comparison}
\end{table}



Since annotators are asked to mask out \textbf{all} objects, \VE{} is much more densely annotated 
compared with existing ones (Table~\ref{tab:dataset_comparison}). Different from YouTube-VOS and DAVIS, we do not specify 
which object to annotate. 
Unlike YouTube-VIS, we do not have a predefined taxonomy for objects. As a result, \VE{} provides 12.29 object annotations per video on average, 
which is 7x more than Youtube-VIS and Youtube-VOS and 4x more than DAVIS. The number of objects per video follows a long-tail distribution shown in Figure~\ref{fig:object_distribution}. In some extreme cases, the number of object instances in a video can be nearly 100 (Figure~\ref{fig:object_distribution}).

\subsection{Dataset annotation}

\textbf{Data source.} We adopt videos from Kinetics-400~\cite{kinetics} for UVO, which contains 10-second 30fps YouTube video clips labeled with human actions. There are several advantages of using Kinetics videos. 
First, Kinetics videos are human-centric and contain diverse sets of human actions and human-object interactions. 
In addition, Kinetics videos are sampled from YouTube with a wide variety of 
different sources: they are taken by both professionals and amateurs, by both cameras and mobile devices, third-person or egocentric. 
Furthermore, Kinetics videos cover very diverse object categories, appearances and motions, including challenging real-world cases 
with occlusion and camera motion.

\textbf{Annotation guideline.}
We acknowledge there is ambiguity in the definition of ``object'', and follow the common approach in the object proposal literature~\cite{PixelObjectness,6544186,Zitnick2014EdgeBL,SelectiveSearch}: objects are defined as things not belonging to background or stuff. We clarify the definition of background or stuff with the annotators through examples: grass, sky, floor, walls, \etc. We note that the difference between objects and stuff have been discussed previously~\cite{heitz2008learning,alexe2010object,caesar2018coco}. On granularity, we ask the annotators to choose 
the coarse end of the object definition: err towards the coarsest possible segmentation that produces meaningful segments.  We identified a few major ambiguities during the annotator training process and address them individually:
\begin{itemize}[noitemsep,topsep=0pt]
    \item Group of objects (connected objects). A group of objects could be marked as one if they stay together through the whole video, such as a stack of bowling balls and a crowd of static people. If an object leaves the group, the object needs to be segmented on its own.
    \item Accessories of humans. An object may have been constantly 
    connected to a human throughout the video. We use a criteria on interaction to decide when to split or merge. For example, in Figure~\ref{fig:compare_vis}, a person working out with a barbell, both the barbell and the sneakers are connected to the person. The person is interacting with the barbell, so it is segmented as an individual object; while the sneakers remain static on the person's feet, so they are annotated as the person.
    \item Objects in the mirror. We make it explicit that mirrored objects by a reflective surface
    are not annotated. 
\end{itemize}

With the guidelines provided, annotators maintained a consistent definition on objects. Since we ask annotators to annotate only masks without object labels, our object masks can be considered as ``anonymous-semantic'' objects. 
Our open-world object segmentation task can be interpreted as grouping pixels/voxels into ``anonymous-semantic'' objects or entities.

\textbf{Annotation frequency.} 
Human annotation of object masks is time consuming. On average, each frame takes 16.3 minutes without linking the object masks over time. To make the annotation more tractable, \VE{} is split into a training set and a test set, where the training set has temporally sparse annotation (1fps) and the test set has dense annotation (30fps).
For the test set, we randomly selects 3-second clips from 1200 videos sampled from Kinetics validation set with considerations to balance each action class. 
This sums up to 108k frames with temporally dense annotations. For the training set, we sample from Kinetics training set and annotate them sparsely at 1fps. We justify these choices through experiments in section~\ref{subsec:baseline_results}.

\textbf{Quality assurance.} We perform a two-stage quality assurance check: frame-level and video-level. On frame-level, auditors (experts) are asked to check whether the mask quality is high, \eg, examining object boundaries. 
Frames with poor annotations found by the auditors are corrected by the annotators.
On video-level, we render each object mask with different colors, which are checked by 3 annotators for temporal consistency. Object masks without majority votes of the 3 annotators are refined again. 

\begin{figure*}
\captionsetup{font=footnotesize}
{\small
\begin{center}
\includegraphics[width=1.0\linewidth]{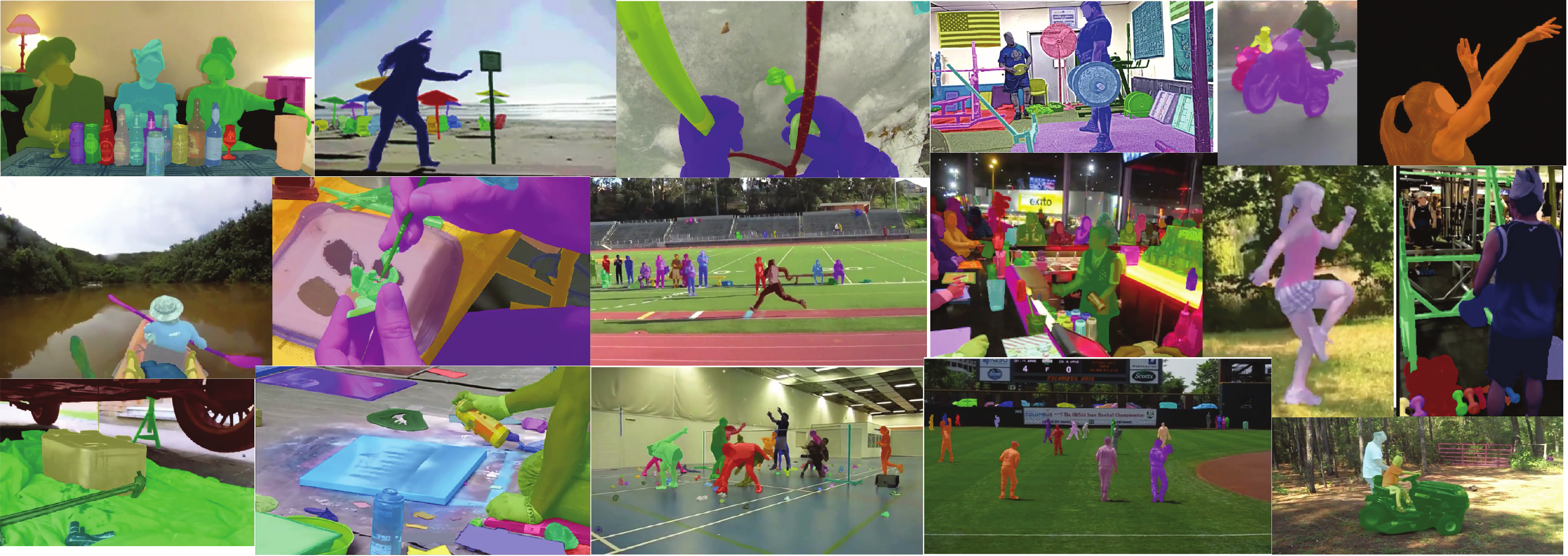}
\vspace{-18pt}
\caption{{\bf Examples of \VE{}}. \VE{} videos are exhaustively annotated with masks regardless of object categories. \VE{} features a wide-range of videos (\eg, third-person/egocentric, professional/amateur, crowded/sparse objects) making it a challenging benchmark. Best viewed in color.}
\label{fig:example_ann}
\end{center}}
\end{figure*}

\subsection{Efficient annotation using mask propagation} \label{subsec:ann_pipeline}

\begin{figure}
\captionsetup{font=footnotesize}
\begin{center}
\includegraphics[width=1.0\linewidth]{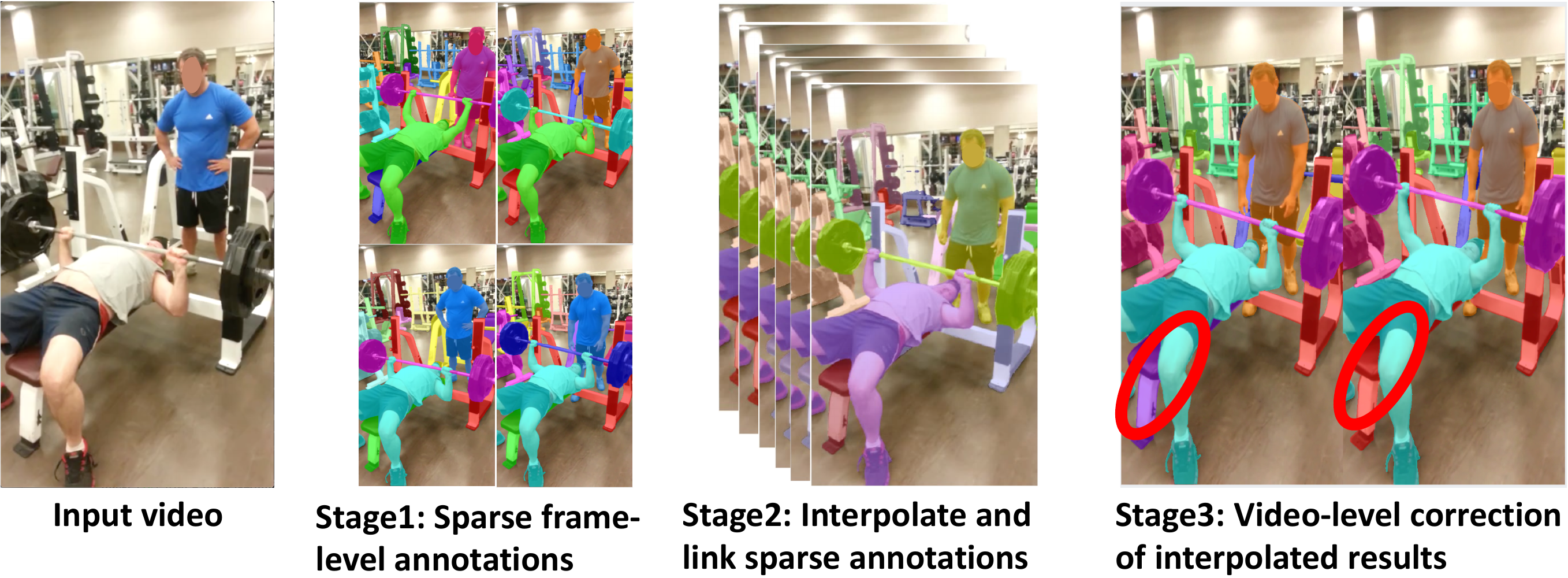}
\vspace{-18pt}
\caption{{\bf Annotation pipeline overview}. We propose a semi-automated pipeline to accelerate the annotation process. We first annotate the videos sparsely (\eg, 1fps), then propagate the masks to the next frames densely. The propagated masks are then corrected by annotators.}
\label{fig:pipeline}
\end{center}
\end{figure}

As shown in Table~\ref{tab:annotation_time}, it takes 45 minutes per frame to annotate all object masks and link them across frames. To speed up the annotation, we detail our proposed semi-automated annotation pipeline (Figure~\ref{fig:pipeline}) in this section. 

{\bf Sparse frame-level segmentation.} Given a video, annotators preview the clip and annotate all the masks at 1 fps. We only ask the annotators to associate each object mask with a unique index and do not require them to link masks over time. This significantly reduces the annotation time from 45 minutes/frame to 16.3 (scratch-frame in Table~\ref{tab:annotation_time}).

\begin{figure}
\captionsetup{font=footnotesize}
\begin{center}
\includegraphics[width=1.0\linewidth]{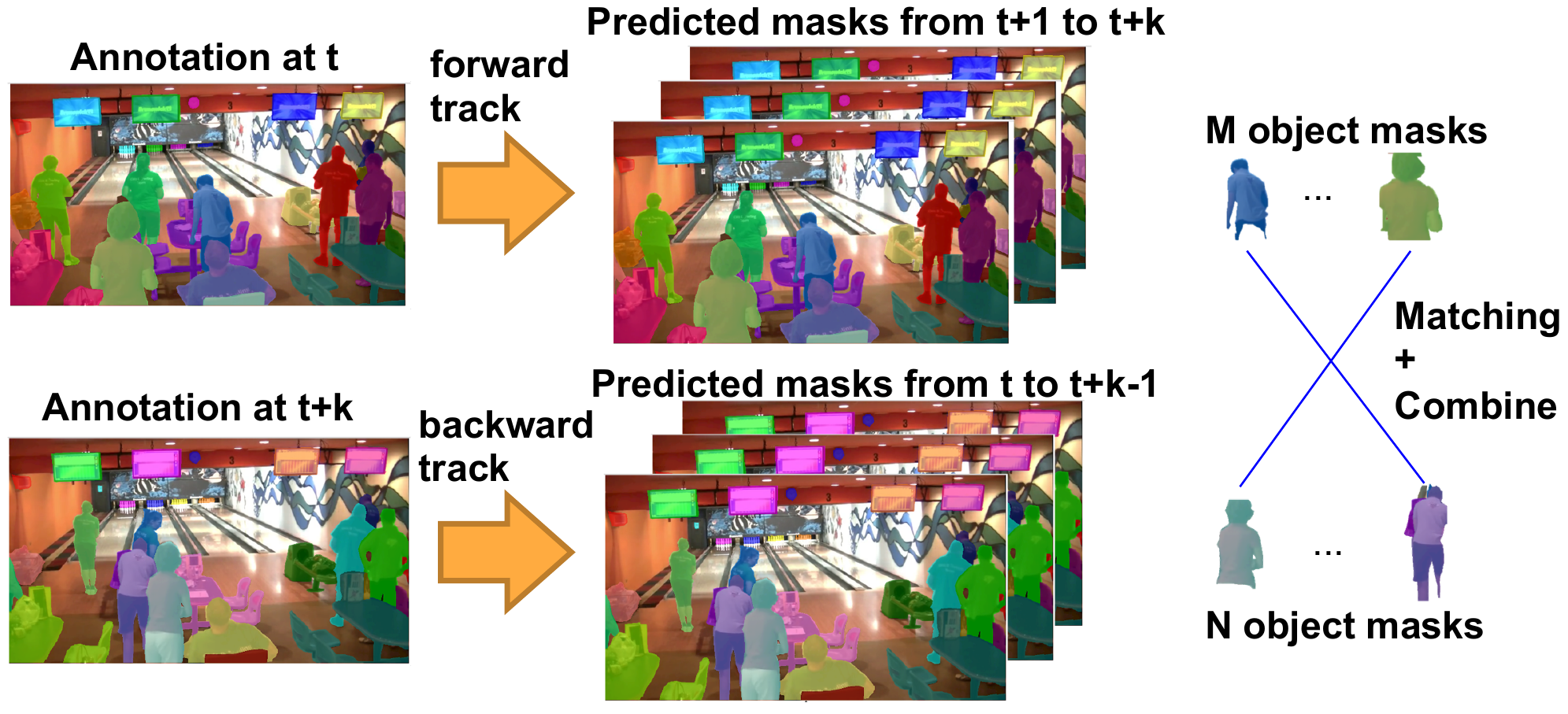}
\vspace{-18pt}
\caption{{\bf Propagating masks from sparsely annotated frames}. 
We interpolate all the object masks between two sparsely annotated frames $t$ and $t+k$. For an un-annotated frame, we generate its annotations by forward and backward tracking the annotations from frame $t$ and $t+k$, matching and combining these two sets of annotations.}
\label{fig:agg_stm}
\end{center}
\end{figure}

{\bf Propagating sparsely annotated masks.} We interpolate the sparsely annotated masks to all the frames. 
To this end, we use Space-Time Memory Network (STM)~\cite{STM} to track the object masks through frames. 
For each un-annotated frame, we consider both forward and backward tracking, \ie, tracking from the closest earlier frame and the closest later frame.
Since objects in sparse annotations may not be linked, forward and backward tracking of the same frame may not match. 
We formulate a maximal bipartite matching problem (\ie, mapping $M$ forward tracked objects to $N$ backward tracked objects) and solve it with Hungarian Matching. Edge costs are a linear combination of overlap (IoU), mask size, object color histogram and object center distance. To combine the forward and backward predictions, we weight their logits by their temporal distances to the un-annotated frame. Our pipeline is summarized in Figure~\ref{fig:agg_stm}.

{\bf Video-level mask correction} We send interpolated masks to annotators for correction. To assist the annotation, we present both the target frame (frame to correct) and the closest frame with sparse annotations. This helps to guide the annotator, ensure temporal consistency and correct linking errors.

The proposed pipeline significantly reduces annotation time  from 45 (naive frame annotation and linking) and 30.7 (copy-paste) to 11 minutes per frame (Table~\ref{tab:annotation_time}).


\begin{table}[]
\captionsetup{font=footnotesize}
    \centering
    {\small
    \begin{tabular}{|c|c|c|c|}
        \hline
        Scratch-frame & Scratch-video & Copy-paste & Our pipeline \\
        \hline
        16.3 & 45.0 & 30.7 & 11.0 \\
        \hline
    \end{tabular}}
    \vspace{-8pt}
    \caption{\textbf{Annotation time per frame (in minutes) in different settings.} Both scratch-frame and scratch-video have no initialization of object masks. Scratch-video requires linking objects across the frames, whereas scratch-frame does not. Copy-paste copies the masks from the sparsely annotated frames as an initialization for dense annotations. With our proposed pipeline, we are able to cut down annotation time by 4x.
    }
    \label{tab:annotation_time}
\end{table}


\subsection{Look into \VE{}}  \label{subsec:data_stats}

\begin{figure}
\captionsetup{font=footnotesize}
\begin{center}
\includegraphics[width=1.0\linewidth]{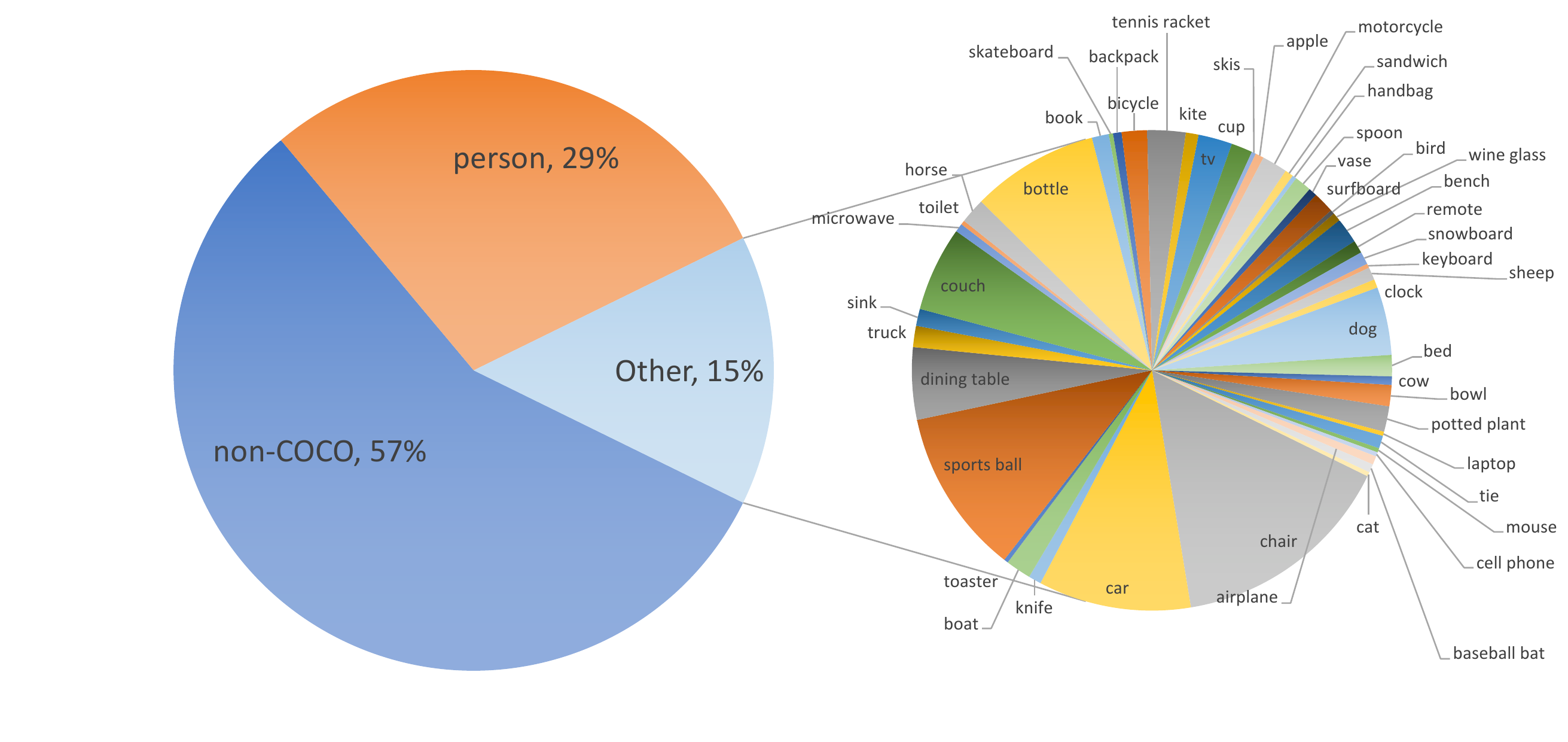}
\vspace{-18pt}
\caption{{\bf Distribution of object categories}. Our dataset contains 57\% objects not belonging to any of the 80 COCO categories. Due to the nature of Kinetics as a human action recognition dataset, our dataset contains 29\% instances labeled as person. Our objects cover 51 COCO categories.}
\label{fig:object_class_distribution}
\end{center}
\end{figure}

\textbf{Open-worldness} Previous datasets use predefined taxonomies, \eg, MSCOCO has 80 object categories. It is natural to ask how well a predefined taxonomy can cover real-world scenarios. 
To understand the open-worldness of \VE{}, we further label the object instances with 80 COCO categories and one additional category to catch all non-COCO classes. 
Figure~\ref{fig:object_class_distribution} presents the distribution of object categories in \VE{}. 
Since Kinetics focuses on human actions, human covers 29~\% of object instances. Besides human, about 15~\% of additional object instances are includes in the COCO taxonomy.
57~\% object instances do not belong to any of the 80 COCO classes. This indicates a limited coverage of a well predefined taxonomy. Many non-COCO objects are less common but are not rare, \eg, ski poles, cable crossover, pliers, dog leashes, pain patch, dried seaweed, \etc. Many of these are not covered by 1.2k LVIS~\cite{8954457} taxonomy neither. There are also many truly ``unknown" objects: one can spot the object but fails to identify what the object is. 

\textbf{Diverse objects and camera motions}. Besides object categories, there are other attributes that can impact the performance of a video object segmentation algorithm.
In the image domain, datasets are often evaluated by dividing objects into different sizes (large, medium and small). For objects in video, motion is an important attribute. We analyze two types of motion: camera motion and object motion. We extract camera motion (rotation and translation) with an off-the-shelf camera pose estimator~\cite{Zhou2017UnsupervisedLO}, and find YTVIS and \VE{} have a similar distributions of camera motions.



\begin{figure*}
\captionsetup{font=footnotesize}
  \begin{subfigure}[b]{0.24\linewidth}
    {\includegraphics[width=\linewidth]{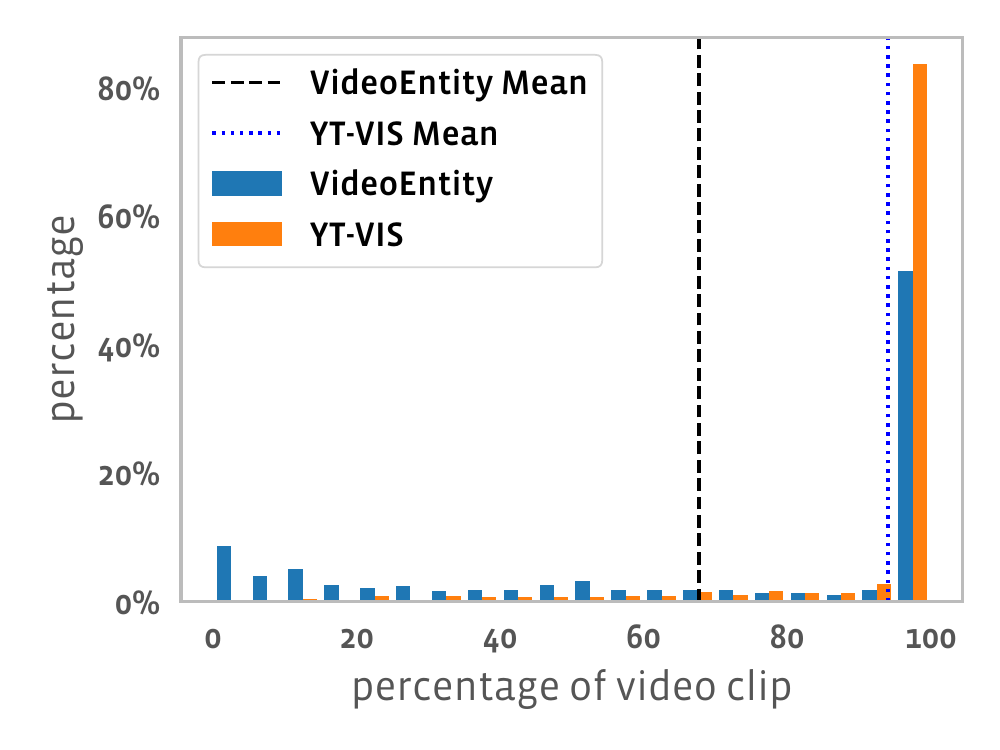}}
    \caption{Object lifespan.}
 \end{subfigure}
 \begin{subfigure}[b]{0.24\linewidth}
    {\includegraphics[width=\linewidth]{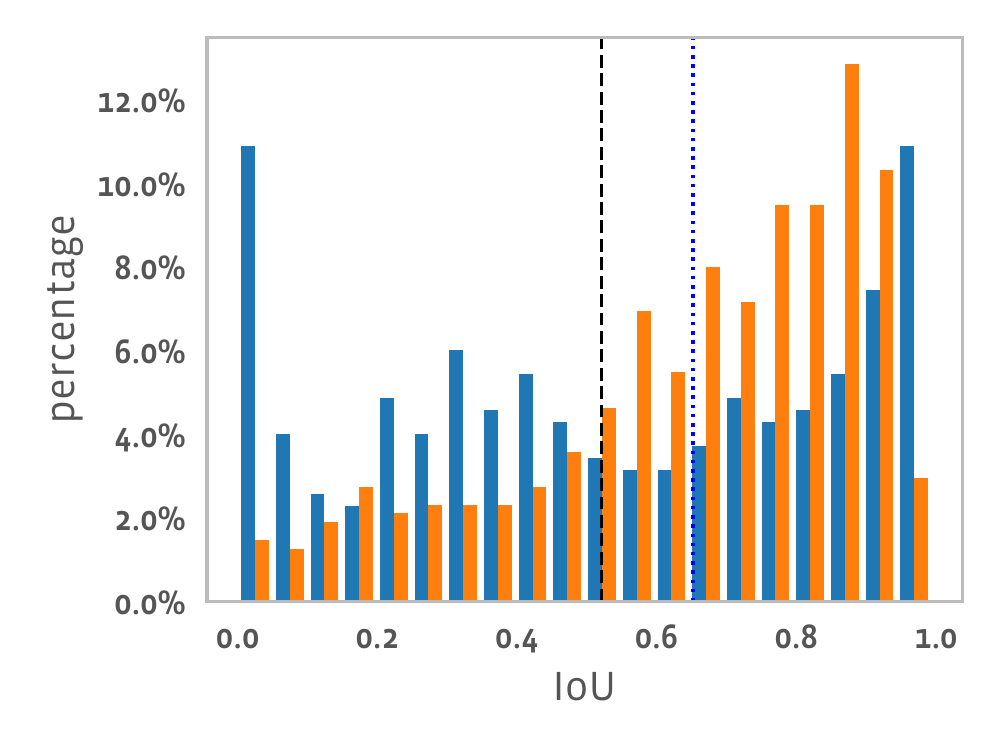}}
    \caption{Object Mask IoU every 5 frames.}
 \end{subfigure}
 \begin{subfigure}[b]{0.24\linewidth}
    {\includegraphics[width=\linewidth]{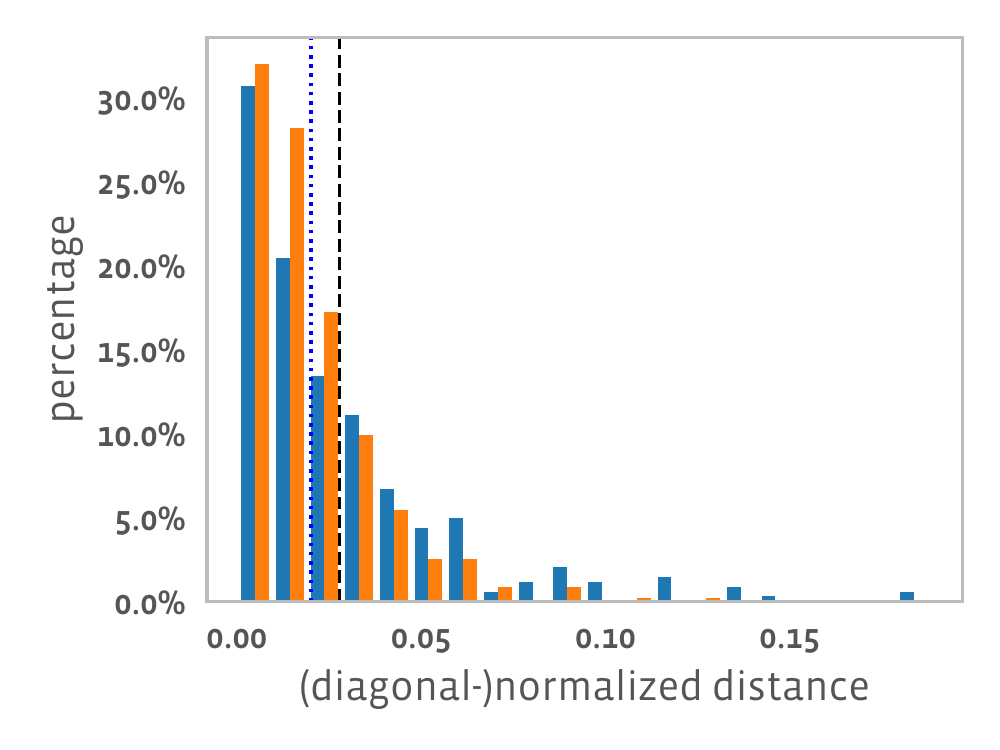}}
    \caption{Object center distance every 5 frames.}
 \end{subfigure}
 \begin{subfigure}[b]{0.24\linewidth}
    {\includegraphics[width=\linewidth]{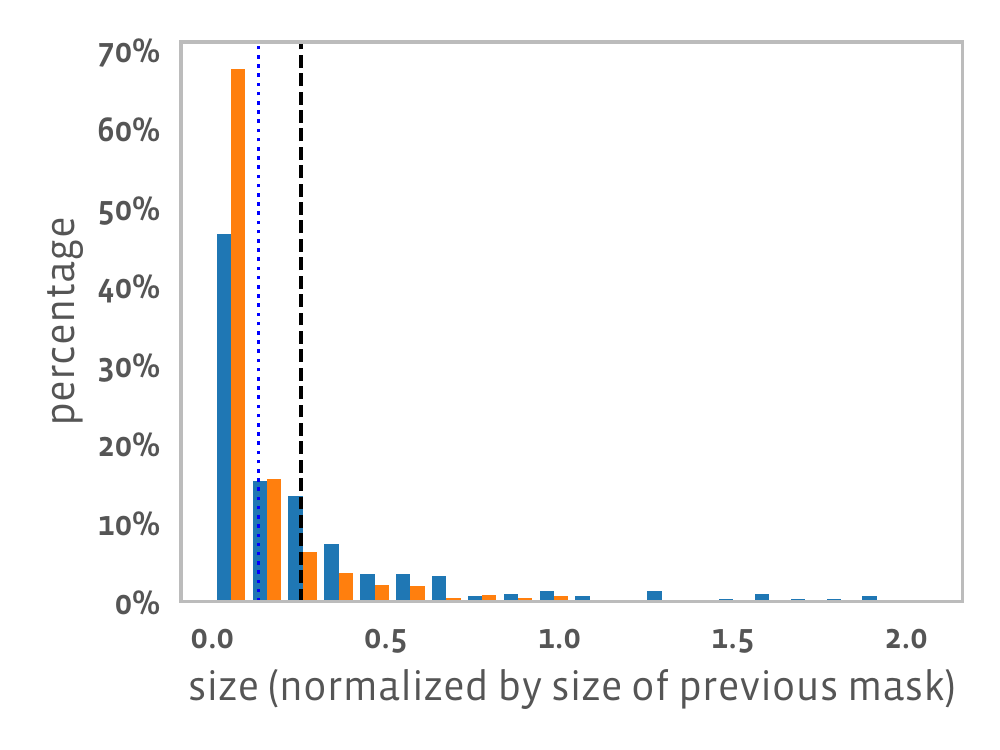}}
    \caption{Object size change every 5 frames.}
 \end{subfigure}
 \caption{\textbf{Comparing object motion statistics with YouTube-VIS.} \VE{} has a wider object lifespan (a), which indicates more frequent object appearing and disappearing. \VE{} objects have a more diverse distributions of motion magnitudes (bcd), also larger motions on average .}
 \label{fig:compare_stats_ytvis}
\end{figure*}

For an object instance, we decouple its motion into two types: disappear/ appear and independent object motion. 
We study the lifespan of an object and compare with YTVIS (Figure~\ref{fig:compare_stats_ytvis}a). \VE{} features a wider distributions of object lifespans (more frequent disappearing and appearing), while YTVIS focuses on objects appeared throughout the entire video. For object motion, we compute pair-wise mask IoU between two timestamps (Figure~\ref{fig:compare_stats_ytvis}b). Our dataset has a broader distribution on maskIoU and on average has smaller IoU (larger motion). We further decouple maskIoU into size-change (\eg, occlusion, shrinking/expanding) and velocity (distance between object mask center). \VE{} provides a larger range of motions and on average has more significant motion, shown in  Figure~\ref{fig:compare_stats_ytvis}c,d. 

\section{Experiments}
\label{sec:baselines}

\subsection{Baselines and implementation details}

\textbf{Frame-level segmentation: Mask R-CNN}~\cite{8237584}. Mask R-CNN is a high-performance two-stage model with an RPN~\cite{FasterRCNN} to generate object proposals and an ROIAlign operation on top of the proposed bounding boxes for classification, box regression, and mask segmentation. Following the common practices, we use COCO17~\cite{Lin2014MicrosoftCC} for pre-training. We also include experiments on LVIS~\cite{8954457}, an expansion of COCO with 1.2k categories.

\textbf{Video-level segmentation: MaskTrack R-CNN}~\cite{YouTubeVIS}. Besides detecting and segmenting objects in each frame, video models also need to correctly link objects across the frames and predict their space-time masks. MaskTrack R-CNN adds an additional tracking head on top of Mask R-CNN, and combines tracking head predictions, class assignments and mask IoU to link objects. We use YouTube-VIS~\cite{YouTubeVIS} (YTVIS) dataset for pre-training and inference. Since YTVIS does not release its validation set annotations and its validation server does not provide class-agnostic evaluation, we split its training data into a training split (1938 videos) and a held-out validation split (300 videos). 

\textbf{Bottom-up segmentation: Hierarchical Graph-Based  Video Segmentation (GBH)}~\cite{GBH}. Besides the aforementioned top-down methods~\cite{8237584,YouTubeVIS}, bottom-up approaches can also be used for unsupervised object segmentation. We adopt the GBH super-voxel algorithm as a baseline. GBH builds a hierarchy of segmentations by progressively grouping similar (super-)voxels. GBH is a non-parametric approach, and we show that in an open-world scenario, it can serve as a competitive baseline. 

\textbf{Object tracking: Space-Time Memory Network (STM)}~\cite{STM}. Tracking (a.k.a semi-supervised video object segmentation) aims at segmenting objects at video level with ground-truth masks provided at the first frame. We adopt the state-of-the-art STM as a baseline on \VE{}. STM uses memory to store features from previous frames and attention to match the current and previous frames for tracking and segmenting objects. 

\textbf{Implementation details}. We use the popular object detection framework Detectron2~\cite{wu2019detectron2} for experiments with class-agnostic Mask R-CNN and MaskTrack R-CNN. We use the open-sourced inference model from~\cite{STM} for the tracking experiments. For super-voxel, we re-implement GBH for better efficiency following the original paper~\cite{GBH} and LIBSVX~\cite{LibSVX}. All models are trained with synchronous SGD with momentum using 8 GPUs. Hyper-parameters follow the settings in the original paper of each dataset (COCO/YTVIS). We use average precision (AP) and average recall (AR) to measure the performance of frame and video level segmentation, and $\mathcal{J}$-score (IoU, region-similarity) and $\mathcal{F}$-score (contour accuracy) for tracking. 


\subsection{Results and analysis} \label{subsec:baseline_results}

\textbf{\VE{} is compatible and complementary with existing datasets}. To show the complementary and compatibility of \VE{} with related instance segmentation datasets, such as COCO and YTVIS, we (pre-)train models on related datasets and optionally finetune on \VE{}, and cross-evaluate on the datasets (Table~\ref{tab:frame_level_cross_eval},\ref{tab:video_level_cross_eval}). On both frame and video level evaluations, \VE{} is more challenging: models trained on COCO or YTVIS suffer from a significant performance drop. Finetuning models on \VE{} offers good gains, but is still lower than the performance of previous datasets. A larger taxonomy (LVIS) also performs poorly on \VE{}, and pre-training on LVIS performs slightly worse than COCO, possibly due to lower performance on ``person'' class (similar results are found in ~\cite{10.1007/978-3-030-58558-7_26}). On video instance segmentation, finetuning without YTVIS gives better results than finetuning with YTVIS, possibly due to the smaller taxonomy and object sparsity in YTVIS.

\VE{} is also compatible with previous datasets. Finetuning on \VE{} offers 1.3\% gain on {AR\footnotesize 100} when evaluating on YTVIS (Table~\ref{tab:video_level_cross_eval}), and only suffers a small gain (3.4\%) on COCO (Table~\ref{tab:frame_level_cross_eval}). Note that \VE{} does not cover 29 out of 80 object categories in COCO. 

On tracking, we evaluate STM on \VE{} and compare with existing datasets, including DAVIS16 (single-object)~\cite{7780454}, DAVIS17~\cite{DAVIS2017} (multi-object) and YouTube-VOS~\cite{YouTubeVOS} (YTVOS)  (Table~\ref{tab:tracking_cross_eval}). \VE{} is slightly harder than existing datasets: performance is lower except for $\mathcal{J}$-score for unseen objects in YTVOS. 

\begin{table}[t]
\captionsetup{font=footnotesize}
  \centering
  {\small
    \begin{tabular}{|c|c|c|c|c|c|}
        \hline
        Train & Test & AP & AP\footnotesize{.5} & AP\footnotesize{.75} & AR\footnotesize{100} \\
        \hline
        \multirow{2}{*}{COCO} & COCO & 37.0 & 63.8 & 38.6 & 49.6 \\
         & \VE{} & 15.9 & 31.0 & 14.2 & 29.7 \\
        \hline
        \multirow{2}{*}{COCO+\VE{}} & COCO & 30.0 & 52.7 & 31.1 & 46.2 \\
         & \VE{} & 22.2 & 42.1 & 20.5 & 41.3 \\
        \hline
        \multirow{2}{*}{LVIS} & LVIS & 15.4 & 26.7 & 15.6 & 37.3 \\
         & \VE{} & 6.8 & 18.1 & 4.1 & 28.3 \\
        \hline
        \multirow{2}{*}{LVIS+\VE{}} & LVIS & 5.7 & 11.3 & 5.1 & 25.4 \\
         & \VE{} & 17.5 & 37.1 & 14.9 & 39.0 \\
        \hline
    \end{tabular}}
    \caption{\textbf{\VE{} frame-level results with Mask R-CNN, cross-evaluated on COCO and LVIS.} As a result of containing a significant amount of objects not in COCO categories, performance of Mask R-CNN trained on COCO drops significantly. Finetuning on \VE{} improves performance significantly, though still much lower than evaluating on COCO only. Models trained on a large-taxonomy (LVIS) also performs poorly on \VE{}, and pretraining with LVIS is slightly worse than COCO.}
    \label{tab:frame_level_cross_eval}
\end{table}


\begin{table}[t]
\captionsetup{font=footnotesize}
  \centering
  {\small
    \begin{tabular}{|c|c|c|c|c|c|}
        \hline
        Train & Test & AP & AP\footnotesize{.5} & AP\footnotesize{.75} & AR\footnotesize{100} \\
        \hline
        \multirow{2}{*}{YTVIS} & YTVIS & 34.7 & 56.6 & 39.1 & 41.0 \\
         & \VE{} & 6.0 & 10.6 & 5.0 & 7.0 \\
        \hline
        \multirow{2}{*}{\makecell{YTVIS\\+\VE{}}} & YTVIS & 25.9 & 43.9 & 27.5 & 42.3 \\
         & \VE{} & 7.0 & 15.1 & 5.1 & 13.2 \\
        \hline
        \multirow{2}{*}{\VE{}} & YTVIS & 13.0 & 27.1 & 12.4 & 23.0 \\
         & \VE{} & 9.3 & 20.9 & 8.2 & 17.2 \\
        \hline
    \end{tabular}}
    \caption{\textbf{\VE{} video-level results with MaskTrack R-CNN, cross-evaluated with YTVIS.} All models are pretrained on COCO. Performance of MaskTrack R-CNN drops significantly when evaluated on \VE{}: there are unseen objects (open-world) and more significant motions (sec~\ref{subsec:data_stats}). A model finetuned on \VE{} can gain when evaluating on YTVIS, but not vice versa.}
    \label{tab:video_level_cross_eval}
\end{table}


\begin{table}[t]
\captionsetup{font=footnotesize}
  \centering
  {\small
  \begin{tabular}{|c|c|c|c|}
    \hline
    Train & Test & $\mathcal{J}$-score & $\mathcal{F}$-score \\
    \hline
    YTVOS+DAVIS & DAVIS16 & 0.887 & 0.899 \\
    \hline
    YTVOS+DAVIS & DAVIS17 & 0.792 & 0.843 \\
    \hline
    \multirow{2}{*}{YTVOS} & YTVOS\footnotesize{seen} & 0.797 & 0.842 \\
     & YTVOS\footnotesize{unseen} & 0.728 & 0.809 \\
    \hline 
    YTVOS+DAVIS & \VE{} & 0.741 & 0.795 \\
    \hline
  \end{tabular}}
  \caption{\textbf{\VE{} tracking results with STM, cross-evaluated with DAVIS and YTVOS.} The overall performance on \VE{} has a slight drop compared to other datasets, and is closer to the unseen scenario in YTVOS.}
  \label{tab:tracking_cross_eval}
\end{table}

\textbf{Open-world detection and segmentation are challenging}. We evaluate the effect of open-world vs. closed-world in detection and segmentation. To simulate an open-world effect on COCO and YTVIS, we follow the common practice~\cite{DeepMask15} by splitting the classes into two sets: classes overlapped with VOC~\cite{PascalVOC} and classes unique to COCO/YTVIS (non-VOC). By training only on VOC classes and test on all or non-VOC classes, we can understand how well a detector performs on unseen (open-world) objects. Specifically, COCO contains all 20 VOC classes, and YTVIS contains 10 VOC classes.

Results are presented in Table~\ref{tab:open_world_result}. When training in closed-world setting and testing in open-world, all models suffer a significant drop in performance. This suggests that existing detectors are bounded to detect only in-taxonomy objects and are not capable of performing open-world detection. In addition, we observe that when enlarging taxonomy (VOC to COCO, COCO to \VE{}), performance on in-taxonomy (seen) objects may decrease slightly. This suggests that open-world detection might involve a per-class performance trade-off when increasing taxonomy size.

\begin{table*}[t]
\captionsetup{font=footnotesize}
  \centering
  {\small
    \begin{tabular}{|c|c|c|c|c|c|c|}
        \hline
        Model & Data & Categories used & All AR\footnotesize{100} & Seen AR\footnotesize{100} & Unseen AR\footnotesize{100} \\
        \hline
        \multirow{2}{*}{Mask R-CNN} & COCO & VOC & 32.8 (-16.8) & 51.3 (+0.3) & 8.8 (-39.4) \\ 
        \cline{2-6}
        & COCO+\VE{} & COCO & 20.5 (-20.8) & 49.5 (+0.8) & 8.9 (-23.6) \\
        \hline
        \multirow{2}{*}{MaskTrack R-CNN} & YTVIS & VOC & 28.8 (-12.2) & 41.4 (+1.2) & 20.3 (-21.8) \\
        \cline{2-6}
        & COCO+\VE{} & COCO & 12.6 (-4.6) & 30.2 (+1.2) & 3.5 (-6.1) \\
        \hline
    \end{tabular}}
    \caption{\textbf{Detecting and segmenting objects in the open-world is more challenging.} Numbers in the bracket indicates difference compared to training on all objects (regardless of class) in the training set. In all settings, training on only a subset of categories decreases Average Recall (AR) significantly for unseen classes/ overall performance, but may slightly improve AR on seen classes.}
  \label{tab:open_world_result}
\end{table*}

\textbf{A possible alternative: bottom-up super-voxel}. Unlike top-down approaches, bottom-up approaches, such as super-voxel algorithms, are by nature taxonomy free. They are typically non-parametric and do not rely on labeled data to train, and therefore, is a natural baseline for open-world problems. We evaluate GBH~\cite{GBH} on \VE{} and YTVIS.

Since GBH provides an over-segmentation of a video, metrics in object detection and segmentation, such as AP/AR, are not directly applicable. On the other hand, super-voxel metrics such as under segmentation error, segmentation accuracy (SA3D) and boundary recall distance~\cite{LibSVX} are not comparable with AP/AR. 

To quantitatively compare bottom-up approach with top-down approach, we adopt an approximate AR for super-voxel algorithms. AR is computed as the average of true-positive detections at multiple IoU thresholds divided by total objects. We make two relaxations. First, we relax the one-to-one mapping in object assignment due to over-segmentation in super-voxels. Specifically, a super-voxel is assigned to the object that has the largest intersection (similar to SA3D); an object, therefore, may be assigned multiple super-voxels. The combination of the assigned super-voxels provides an object-level prediction. Second, with no ranking score, it is not possible to take a cut-off using a fixed number of proposals for each video (\eg, {AR\footnotesize 100} allows maximal 100 proposals per video). We compute {AR\footnotesize k}, where k is the average number of super-voxels for a dataset. The goal of this comparison is not to suggest super-voxel algorithms are stronger or weaker compared with top-down methods (\eg, MaskTrack R-CNN); but rather to provide a possible alternative baseline for open-world problems.

\begin{figure}
\captionsetup{font=footnotesize}
{\small
\begin{center}
\includegraphics[width=0.8\linewidth]{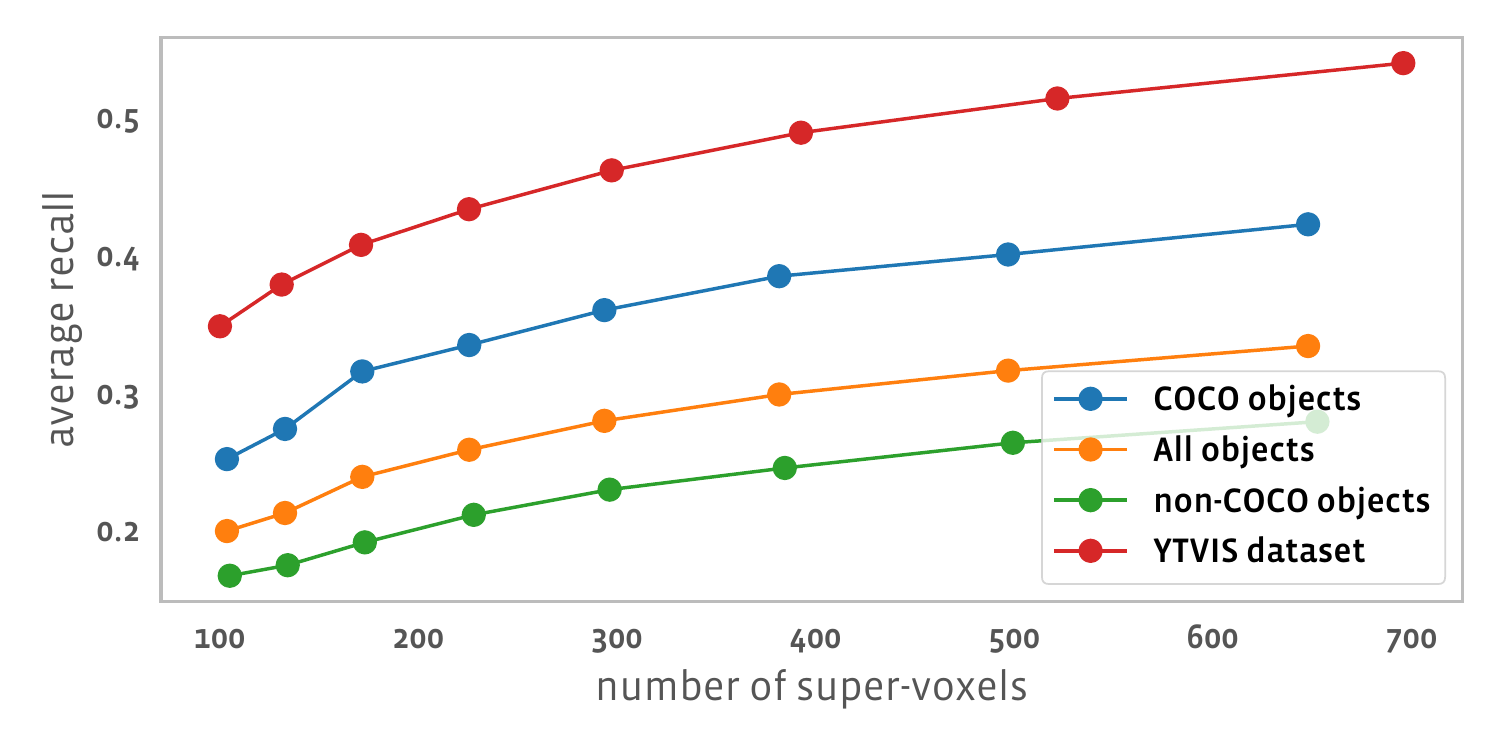}
\vspace{-8pt}
\caption{{\bf GBH performance on \VE{} (all, COCO, non-COCO) and YTVIS}. At an average of 103 super-voxel proposals, approximate AR on \VE{} is 19.8\%, 2.6\% higher than AR{\scriptsize 100} of top-down approach MaskTrack R-CNN. In addition, the gap between unknown and COCO objects is much smaller compared to top-down approach (8.4\% vs. 19.4\%). Even on COCO objects, bottom-up approach is only 4\% behind. For YTVIS, at an average 100 proposals, approximate AR is 34.7\%, 5.9\% higher than only training with VOC objects and 6.3\% lower than training on all objects.}
\label{fig:gbh_videoentity}
\end{center}}
\end{figure}

On \VE{}, super-voxel algorithms achieve 19.8\% for approximate {AR\footnotesize 103}, 2.6\% higher than top-down MaskTrack R-CNN {AR\footnotesize 100} (Figure~\ref{fig:gbh_videoentity}). The gap between non-COCO objects to COCO objects is much smaller than MaskTrack R-CNN. We note that the metric used in top-down and bottom up is not directly comparable since multiple super-voxels can be assigned to one detection. On YTVIS, GBH achieves 34.7\% for approximate {AR\footnotesize 100}, 5.9\% higher than open-world YTVIS (train on VOC classes and test on all) and 6.3\% lower than closed-world YTVIS (Table~\ref{tab:open_world_result}).


\textbf{Moving objects are harder}. In image object segmentation such as COCO, objects are divided into multiple groups based on size, so algorithms are evaluated on how well they perform on large/ medium/ small objects. For video-level evaluation, object motion is another important attribute to look at. We divide the objects by their motion attributes: mask-IoU, object velocity, and object size change. For each attribute, we divide objects into 3 equally-sized groups: high-motion, mid-motion, and low-motion. For example, object with low mask-IoU, high velocity, and large size change is considered as high-motion. 

Results are shown in Table~\ref{tab:object_motion_compare}. Objects with higher motion have worse performance on video models for both \VE{} and YTVIS. In particular, when motion is quite significant ({AR\footnotesize high}), performance drops significantly. This suggests that current models may fail to handle significant motion in objects. In addition to object motions, we also found that larger camera-motion could lead to inferior results, on both YTVIS and \VE{} (refer to supplementary materials).

\begin{table}[t]
\captionsetup{font=footnotesize}
  \centering
  {\small
  \begin{tabular}{|c|c|c|c|c|}
    \hline
    Dataset & Attribute & AR\footnotesize low & AR\footnotesize mid & AR\footnotesize high \\
    \hline
    \multirow{3}{*}{\VE{}} & IoU & 30.5 & 18.7 & 2.6 \\
    \cline{2-5}
     & size change & 32.1 & 16.5 & 3.9 \\ \cline{2-5}
     & velocity & 29.3 & 16.1 & 7.4 \\
    \hline
    \multirow{3}{*}{YTVIS} & IoU & 60.3 & 41.6 & 20.9 \\ \cline{2-5}
     & size change & 61.0 & 41.2 & 20.9 \\ \cline{2-5}
     & velocity & 49.8 & 44.9 & 28.4 \\
     \hline
  \end{tabular}}
  \caption{\textbf{Object with larger motion is harder.} Performance is measured by {AR\footnotesize{100}}. Low, mid and high indicates the magnitude of change with respect to the three attributes. A low inter-frame IoU indicates larger change of the object in time, and we see that performance is consistently lower on \VE{} and YTVIS. We decouple inter-frame IoU into size change and object velocity, and observe that performance is more distinguishable with respect to size change.}
  \label{tab:object_motion_compare}
\end{table}


\textbf{Is temporally dense annotation necessary?} Temporally dense annotations (30fps) are costly to obtain and YTVIS circumvents the problem by annotating sparsely (6fps). We examine the choice in the context of \VE{} by training and evaluating on a down-sampled version at 6fps. Models trained at 6fps and 30fps have only a minor difference evaluating on 6fps, but a 2.7\% gap on {AR \footnotesize 100} evaluating at 30fps (Table~\ref{tab:data_temporal_sparsity}): temporally dense segmentation benefits from temporally dense annotations in evaluation.

For training, from a cost-efficient perspective, we may potentially trade-off high fps annotation with lower fps to scale for more videos. We examine this choice by using 1fps tracking-free annotation (no linking of objects) with intermediate data generated by STM~\cite{STM}. This mirrors the mask propagation step in our annotation pipeline (sec~\ref{subsec:ann_pipeline}). Training on the same number of videos with 1fps plus interpolated masks, the model is able to achieve similar performance compared to 6fps annotated data (Table~\ref{tab:data_temporal_sparsity}). By trading-off sparsity for more videos and using 2x number of videos in training data, we are able to further closen the gap with 30fps annotations: sparse annotation could be sufficient for training.

\begin{table}[t]
\captionsetup{font=footnotesize}
  \centering
  {\small
  \begin{tabular}{|c|c|c|c|}
    \hline
    Train data & Test data & AP & AR\footnotesize{100} \\
    \hline
    \multirow{2}{*}{30fps} & 30fps & 9.3 & 17.2 \\
     & 6fps & 6.7 & 14.1 \\
    \hline
    \multirow{2}{*}{6fps} & 30fps & 7.2 & 14.5 \\
     & 6fps & 6.6 & 14.8 \\
    \hline
    1fps + interpolation & 30fps & 7.2 & 15.0 \\
    \hline
    1fps + interplt. + 2x data & 30fps & 8.4 & 16.0 \\
    \hline
  \end{tabular}}
  \caption{\textbf{Evaluating temporally dense video segmentation requires dense annotations; training with sparsely annotation can also achieve competitive performances.} A sparsely annotated evaluation data cannot tell the difference between the first two models, while there is a 2.7\% gap on AR when test on densely annotated data. On the other hand, for training, training with 1fps groundtruth plus interpolated masks for intermediate frames gives competitive performance compared to 6fps. By trading-off annotation density and annotating additional data at 1fps, we can further improve the performance and closer the gap with 30fps data.}
  \label{tab:data_temporal_sparsity}
\end{table}

\section{Conclusion}
\label{sec:discussion}

We have presented \VE{}, a new benchmark for open-world object segmentation--an unsolved challenging problem with various real-world applications. Compared with current benchmarks, \VE{} is not only different in the open-world problem setup but also multiple times larger in terms of size and annotations. We believe that \VE{} will enable more comprehensive video understanding research such as long-term video modeling and complex video understanding tasks. 

We plan to release the benchmark, continue annotating more videos, provide a test server, and host a challenge workshop for open-world object segmentation at ICCV'21.

\textbf{Acknowledgement.} The authors would like to thank Mike Zheng Shou, Abhijit Ogale, Dhruv Mahajan, Kristen Grauman, Lorenzo Torresani, Rakesh Ranjan and Federico Perazzi for their valuable feedback on the dataset, Haoqi Fan and William Wen for engineering supports. 


{\small
\bibliographystyle{ieee_fullname}
\bibliography{egbib}
}

\clearpage



\appendix
\section*{Appendix}

\section{Camera Motion Statistics and Ablations}

\begin{figure}
\captionsetup{font=footnotesize}
  \begin{subfigure}[b]{1.0\linewidth}
    {\includegraphics[width=\linewidth]{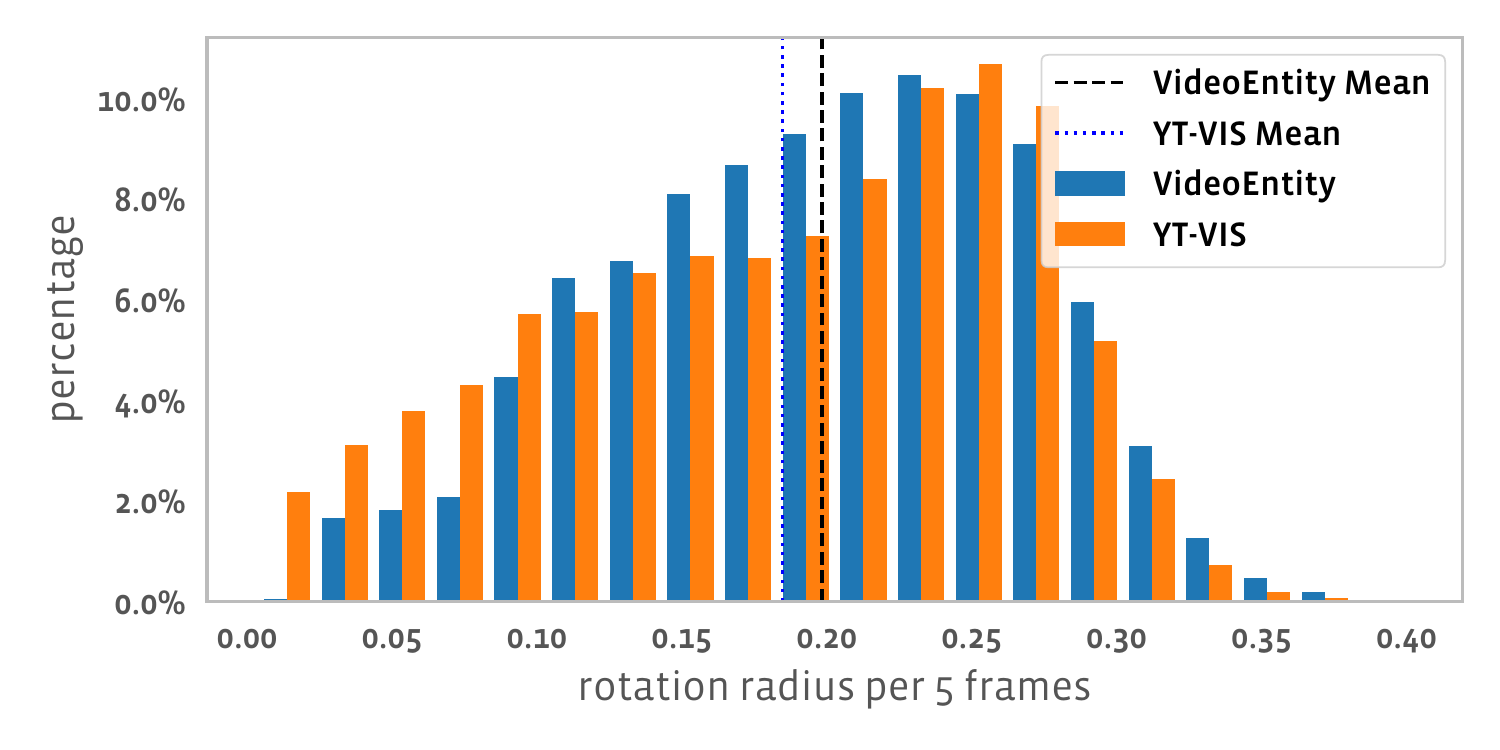}}
    \caption{Camera rotations.}
 \end{subfigure}
 \begin{subfigure}[b]{1.0\linewidth}
    {\includegraphics[width=\linewidth]{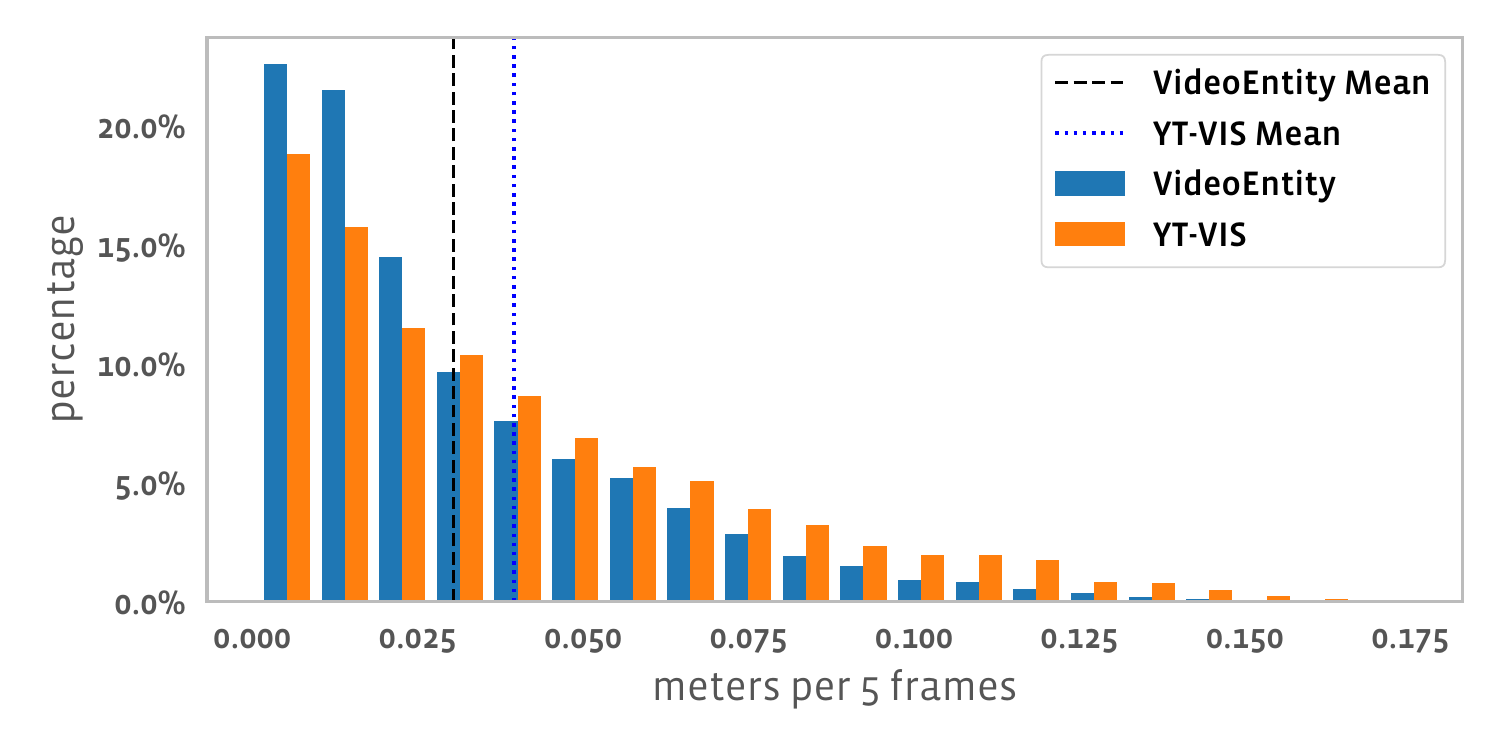}}
    \caption{Camera translations.}
 \end{subfigure}
 \caption{\textbf{Comparing video camera motion statistics between \VE{} and YouTube-VIS.} For video camera motion, YTVIS and Video-Entity follows similar distributions (a and b).}
 \label{fig:compare_camera_motion}
\end{figure}

We use an off-the-shelf camera pose estimator~\cite{Zhou2017UnsupervisedLO} to compute camera motion for \VE{} and YouTube-VIS (YTVIS). Distributions are shown in Fig.~\ref{fig:compare_camera_motion}. Both datasets offer a wide range of camera translations and rotations: \VE{} has slightly higher camera rotations and YTVIS has higher camera translations on average.

{\bf Open-world segmentation is harder on larger camera motion videos}. We also study the impact of two types of camera motions (rotation and translation) on our open-world segmentation task. For each motion type, we divide the videos into two subsets: high motion and low motion. Results are shown in Table~\ref{tab:camera_motion_compare}. ON both YTVIS and \VE{}, MaskTrack R-CNN performs worse on videos with higher camera motion.

{\bf Pre-training ImageNet provides a better initialization}. Existing instance segmentation models, such as Mask R-CNN, are often pretrained on ImageNet~\cite{deng2009imagenet}. Since our dataset is collected from Kinetics400~\cite{kinetics} videos, it is natural to ask if domain differences matter (from image to video) and pretraining on Kinetics400 is more suitable. We replace the ImageNet pretraining with Kinetics400 pretraining (a frame-based model) for both Mask R-CNN and MaskTrack R-CNN. 

\begin{table}
\captionsetup{font=footnotesize}
  \centering
  {\small
  \begin{tabular}{|c|c|c|c|}
    \hline
    Dataset & Attribute & AR\footnotesize low & AR\footnotesize high \\
    \hline
    \multirow{2}{*}{\VE{}} & camera rotation & 19.0 & 15.3 \\
    \cline{2-4}
     & camera translation & 18.5 & 15.9 \\ 
    \hline
    \multirow{2}{*}{YTVIS} & camera rotation & 46.1 & 36.1 \\ \cline{2-4}
     & camera translation & 44.6 & 37.6 \\ 
     \hline
  \end{tabular}}
  \caption{\textbf{Open-world segmentation is harder on larger camera motion videos.} Performance is measured by {AR\footnotesize{100}}. Lower translation/ rotation in camera pose has higher performance compared to larger translation/ rotation on both \VE{} and YTVIS.}
  \label{tab:camera_motion_compare}
\end{table}

Results are summarized in Table~\ref{tab:kinetics_pretrain}. When replacing ImageNet with Kinetics for the first stage of pre-training, all results got worse. One possible cause is that Kinetics videos are labeled by human actions (such as shooting soccer ball), and are not usually object-centered. As a result, Kinetics taxonomy and videos may not be proper for pre-training object detectors compared to ImageNet.

\begin{table}
\captionsetup{font=footnotesize}
  \centering
  {\small
  \begin{tabular}{|c|c|c|c|}
    \hline
    Dataset & Pre-train & AP & AR\footnotesize 100 \\
    \hline
    \multirow{2}{*}{COCO} & ImageNet & 37.0 & 49.6 \\
    \cline{2-4}
     & Kinetics & 30.0 & 44.4 \\ 
    \hline
    \multirow{2}{*}{\VE{}-Frame} & ImageNet & 22.2 & 41.3 \\
    \cline{2-4}
     & Kinetics & 18.4 & 37.3 \\ 
    \hline
    \multirow{2}{*}{\VE{}-Video} & ImageNet & 9.3 & 23.0 \\
    \cline{2-4}
     & Kinetics & 7.5 & 14.0 \\ 
     \hline
  \end{tabular}}
  \caption{\textbf{Pre-training on ImageNet is better than on Kinetics}. This indicates the taxonomy of ImageNet provides a better initialization for open-world object segmentation.}
  \label{tab:kinetics_pretrain}
\end{table}

\section{Implementation Details}
When finetuning on \VE{} for Mask R-CNN and MaskTrack R-CNN, we search hyper-parameters on total number of iterations/ epochs and initial learning rate. For learning schedule, we follow the existing settings in~\cite{wu2019detectron2} and \cite{YouTubeVIS}. For finetuning Mask R-CNN, we use an initial learning rate of $5 \times 10^{-4}$ and train for 4.5k iterations; for finetuning MaskTrack R-CNN, we use an initial learning rate of $5 \times 10^{-4}$ for $3240$ iterations. Dataset interpolation pipeline is implemented in PyTorchVideo~\cite{fan20201pytorchvideo}.

\end{document}